%% file: main.tex
\documentclass[10pt,twocolumn,letterpaper]{article}

\usepackage[pagenumbers]{iccv} 

\input{preamble}

\definecolor{iccvblue}{rgb}{0.21,0.49,0.74}
\usepackage[pagebackref,breaklinks,colorlinks,allcolors=iccvblue]{hyperref}

\def\paperID{979} 
\def\confName{ICCV}
\def\confYear{2025}

\title{VLDrive: Vision-Augmented Lightweight MLLMs for Efficient Language-grounded Autonomous Driving
}

\author{ 
    Ruifei Zhang$^{1,2}$, 
    Wei Zhang$^{4}$, 
    Xiao Tan$^{4}$, 
    Sibei Yang$^{3}$, 
    Xiang Wan$^{2}$, 
    Xiaonan Luo$^{5}$, 
    Guanbin Li$^{3,6}$\thanks{Corresponding author: liguanbin@mail.sysu.edu.cn} \\
    $^1$ The Chinese University of Hong Kong, Shenzhen 
    $^2$ Shenzhen Research Institute of Big Data \\
    $^3$ Sun Yat-sen University 
    $^4$ Baidu Inc. 
    $^5$ Guilin University of Electronic Technology \\
    $^6$ Guangdong Key Laboratory of Big Data Analysis and Processing \\
}

\begin{document}
\maketitle
\input{sec/0_abstract}    
\input{sec/1_intro}
\input{sec/2_related}

\input{sec/3_method}

\input{sec/4_experiments}
\input{sec/5_conclusion}

\section*{Acknowledgments}
This work is supported in part by the National Key R\&D Program of China (2024YFB3908503), in part by the National Natural Science Foundation of China (62322608), in part by the Shenzhen Longgang District Science and Technology Innovation Special Fund (No. LGKCYLWS2023018), in part by the Futian Healthcare Research Project (No.FTWS002), and in part by the Shenzhen Medical Research Fund (No. C2401036). This work is also sponsored by CIE-Tencent Robotics X Rhino-Bird Focused Research Program.

{
    \small
    \bibliographystyle{ieeenat_fullname}
    \bibliography{main}
}

\end{document}

%% file: preamble.tex
\usepackage{multirow}
\usepackage{graphicx}
\usepackage{bbding}
\usepackage{amsmath}
\usepackage{amssymb}
\usepackage{subcaption,booktabs}
\usepackage{xcolor,colortbl}
\definecolor{lavender}{RGB}{150,123,182}
\definecolor{coral}{RGB}{255,127,80}
\definecolor{teal}{RGB}{0,128,128}
\definecolor{lightred}{RGB}{255,204,203}
\usepackage{pifont}

%% file: sec/0_abstract.tex
\begin{abstract}
Recent advancements in language-grounded autonomous driving have been significantly promoted by the sophisticated cognition and reasoning capabilities of large language models (LLMs). However, current LLM-based approaches encounter critical challenges: (1) Failure analysis reveals that frequent collisions and obstructions, stemming from limitations in visual representations, remain primary obstacles to robust driving performance. (2) The substantial parameters of LLMs pose considerable deployment hurdles. To address these limitations, we introduce VLDrive, a novel approach featuring a lightweight MLLM architecture with enhanced vision components. VLDrive achieves compact visual tokens through innovative strategies, including cycle-consistent dynamic visual pruning and memory-enhanced feature aggregation. Furthermore, we propose a distance-decoupled instruction attention mechanism to improve joint visual-linguistic feature learning, particularly for long-range visual tokens. Extensive experiments conducted in the CARLA simulator demonstrate VLDrive's effectiveness. Notably, VLDrive achieves state-of-the-art driving performance while reducing parameters by 81\% (from 7B to 1.3B), yielding substantial driving score improvements of \textbf{15.4}\%, \textbf{16.8}\%, and \textbf{7.6}\% at tiny, short, and long distances, respectively, in closed-loop evaluations. Code is available at  \url{https://github.com/ReaFly/VLDrive}.
\end{abstract}

%% file: sec/1_intro.tex
\section{Introduction}
\label{sec:intro}

Autonomous driving has experienced significant development in recent years, evolving from modular and segregated design to integrated end-to-end networks~\cite{liang2020pnpnet,zhang2024offsetnet,sadat2020perceive,hu2023planning,shao2023safety}. However, these networks still rely on specific inputs such as target points or action commands to guide their driving behaviors, significantly constraining the interaction with humans and applicability in real-world scenarios.

\begin{figure}[t]
\centering
\includegraphics[width=\linewidth, trim=0 0 0 0]{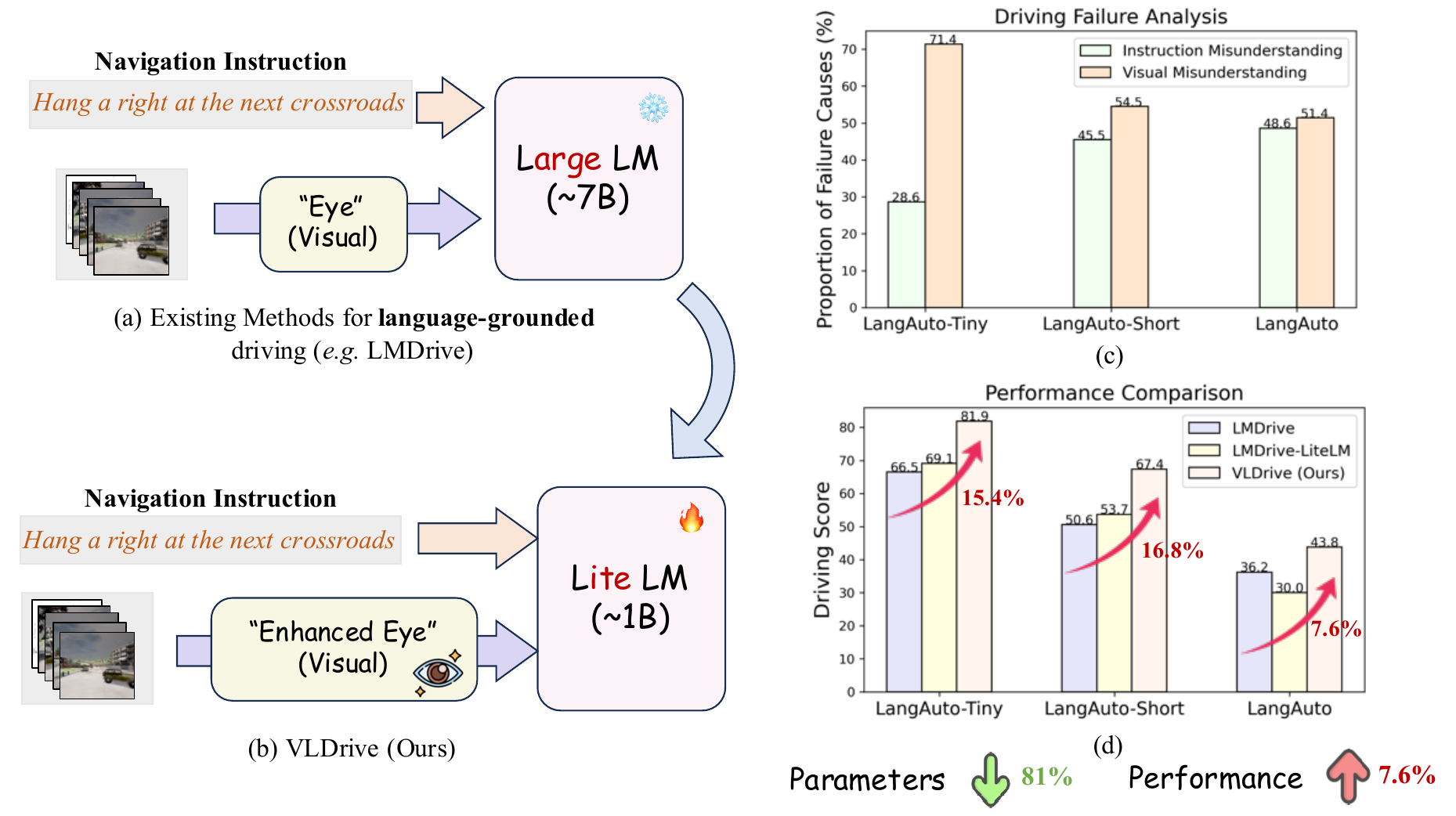}
\caption{(a) Existing methods for language-grounded driving. (b) Our proposed VLDrive: a novel framework featuring a lightweight MLLM architecture with enhanced vision components. (c) Driving failure analysis of existing methods based on three evaluation runs. (d) Performance comparison between VLDrive and both versions of LMDrive, highlighting our method's superior driving performance with fewer parameters.}
\label{fig:first}
\end{figure}

The emergence of large language models (LLMs)~\cite{touvron2023llama,zheng2024judging,zhu2023minigpt,liu2024visual} has catalyzed a revolution in autonomous driving. Impressed by their advanced cognition and logical reasoning capabilities, a multitude of studies have integrated LLMs into autonomous driving systems and achieved noteworthy results~\cite{xu2023drivegpt4,shao2023lmdrive,wang2023drivemlm,wang2024drivecot}. Among them, LMDrive~\cite{shao2023lmdrive} successfully achieves human-friendly \textbf{language-grounded} autonomous driving, where vehicle behaviors are \textbf{solely} guided by natural language instructions (see Fig. 1(a)), significantly enhancing human-vehicle interaction. 

However, despite remarkable advancements, these approaches encounter significant practical deployment challenges, characterized by suboptimal driving performance and excessive computational parameters. To address these issues, we conduct a thorough investigation of driving failure cases in current approaches, coupled with performance analysis across different LLM scales, yielding two critical insights: 1) {\textit{\textbf{Deficiencies in visual comprehension are the predominant cause of driving failures.}}} We analyze the failure cases of existing language-grounded autonomous driving models and identify two primary causes: \textbf{instruction misunderstanding}, manifested as the agent deviating from the route; and \textbf{visual misunderstanding}, characterized by failures due to collisions or blocks involving vehicles or layouts. The results in Fig.~\ref{fig:first}(c) reveal that limitations in visual comprehension frequently emerge as a critical factor contributing to driving task failures.
2) {\textit{\textbf{Driving performance does not scale linearly with LLM's parameter count}}}. Through comprehensive evaluations across three benchmarks, we compare the full-scale LMDrive (7B parameters) with its lightweight counterpart, LMDrive-LiteLM (1.3B trainable parameters). As demonstrated in Fig.~\ref{fig:first}(d), LMDrive-LiteLM achieves comparable driving performance despite its significantly reduced parameter count. These findings strongly suggest that massive parameter scales in LLMs may not be a critical prerequisite for achieving superior driving performance. The insights derived from these analyses illuminate the path toward developing a parameter-efficient yet high-performing alternative for language-grounded driving applications.

Building on these considerations, in our work, we propose \textbf{VLDrive}, a vision-augmented lightweight language model for efficient language-grounded autonomous driving, as showcased in Fig.~\ref{fig:first}(b). We enhance the model's visual comprehension by drawing inspiration from human driving competencies, focusing on two key aspects: prioritized attention allocation to sparse but critical agents and objects, and temporal trajectory memory for behavioral sequence prediction. These enhancements are achieved through our proposed Cycle-consistent Dynamic Visual Pruning (CCDP) and Memory-enhanced Feature Aggregation (MEFA) mechanisms. Specifically, 
CCDP framework consists of token sparsification and training-only token reconstruction. The sparsification mechanism enables adaptive selection of critical visual tokens and dynamic adjustment of token count based on driving scenario complexity. The reconstruction process preserves information integrity by explicitly recovering features of pruned tokens, enabling robust visual perception while minimizing computational overhead. Additionally, MEFA incorporates a memory bank of adjacent frames to capture temporal dependencies, providing critical temporal cues for visual feature enhancement.

Furthermore, we introduce a Distance-decoupled Instruction Attention (DDIA) mechanism to enhance instruction comprehension capabilities of lightweight language models. This design stems from our observation that increasing numbers of long-range visual tokens may dilute instruction attention (see Fig.~\ref{fig:atts}), potentially leading to \textbf{instruction misunderstanding}. DDIA alleviates this issue and enhances joint visual-linguistic feature learning and alignment, resulting in more accurate trajectory planning.
Extensive closed-loop experiments conducted on the CARLA~\cite{dosovitskiy2017carla} simulation platform demonstrate the effectiveness of our proposed method, surpassing state-of-the-art approaches by \textbf{15.4}\%, \textbf{16.8}\% and \textbf{7.6}\% in driving scores at tiny, short and long distances, respectively.

To summarize, our contributions are as follows:
\begin{itemize}
    \item We identify that visual comprehension deficiencies constitute the primary bottleneck in language-grounded driving, with performance scaling non-linearly with language model size. This observation motivates our proposed paradigm: a lightweight language model enhanced by vision-augmented strategies, which simultaneously reduces computational overhead and improves driving performance, facilitating practical deployment.
    \item We propose VLDrive, a lightweight architecture enhanced with vision-centric strategies for language-grounded driving. The framework incorporates CCDP for adaptive visual signal extraction and MEFA for temporal information integration. Additionally, DDIA facilitates visual-language alignment for robust navigation instruction following. These complementary strategies collectively enable more reliable autonomous driving.
    \item We evaluate our approach through closed-loop simulation experiments on standard language-grounded driving benchmarks using the CARLA platform. VLDrive achieves state-of-the-art performance while using significantly fewer parameters.  
\end{itemize}

%% file: sec/2_related.tex
\section{Related Work}
\subsection{End-to-End Autonomous Driving}
Early end-to-end autonomous driving methods can be mainly divided into two categories: imitation learning~\cite{codevilla2019exploring,zhang2017query,hawke2020urban} and reinforcement learning~\cite{liang2018cirl,toromanoff2020end,chekroun2023gri,zhang2021end}. Subsequently, benefiting from the robust feature modeling capabilities of the Transformer~\cite{vaswani2017attention}, 
researchers introduced additional modalities to provide complementary information for autonomous driving~\cite{prakash2021multi,shao2023safety,zhang2024interactive}. For example, Transfuse~\cite{prakash2021multi} uses Transformer to merge features from RGB and LiDAR bird's eye view (BEV) images to enhance the global comprehension of 3D scenes. InterFuse~\cite{shao2023safety} improves the security and interpretability of the model by directly presenting its intermediate features and limiting actions to specified safe sets. In recent years, modular end-to-end planning approaches~\cite{hu2023planning,sadat2020perceive,casas2021mp3} have gained increased attention, where all components are connected and jointly optimized towards the ultimate goal, further improving the driving performance.

\subsection{LLMs in Autonomous Driving}
As LLMs continue to evolve and flourish, an increasing number of researchers are incorporating them into autonomous driving tasks, broadening the scope and capabilities of these systems. Among them, a notable strand of research~\cite{marcu2023lingoqa,nie2023reason2drive,sima2023drivelm} employs LLMs for driving-related visual question answering, providing enhanced interpretability for driving decisions. Another line of work seeks to fully exploit the powerful logical reasoning capabilities of LLMs to fundamentally transform the research paradigm in autonomous driving tasks~\cite{mao2023gpt,mao2023language,fu2024drive,cui2024drive}. Specifically, GPT-Driver~\cite{mao2023gpt} feeds LLM with observations and ego-states as language prompts and generates a planned trajectory and corresponding decision-making process in a natural language format. Agent-Driver~\cite{mao2023language} builds an LLM-powered agent, capitalizing on a tool library, a cognitive memory, and a reasoning engine for human-like autonomous driving. Recently, LLMs have been extended to visual modality successfully, making it possible for LLMs to understand the visual information of autonomous driving. \citet{chen2023driving} introduces a multimodal LLM to process vectorized numeric modalities data and produce driving-related question answers as well as action decisions. 
DriveGPT4~\cite{xu2023drivegpt4} curates a visual instruction tuning dataset for interpretable autonomous driving, taming LLM to generate text responses and low-level control signals. Further, based on the CARLA simulation platform, many studies~\cite{shao2023lmdrive,wang2023drivemlm,wang2024drivecot} delve deeper and perform closed-loop evaluation driven by LLMs directly. For instance, LMDrive~\cite{shao2023lmdrive} takes multi-modality data and navigation instructions as input, achieving language-grounded driving and enhanced interaction with humans. Despite the impressive performance powered by LLMs, their considerably heavy parameters pose significant deployment challenges. In this work, we demonstrate that a lightweight language model can effectively respond to instructions and execute driving tasks through efficient visual representation extraction and strengthened vision-language alignment.

\begin{figure*}[t]
\centering
\includegraphics[width=0.71\textwidth, trim=0 0 0 0]{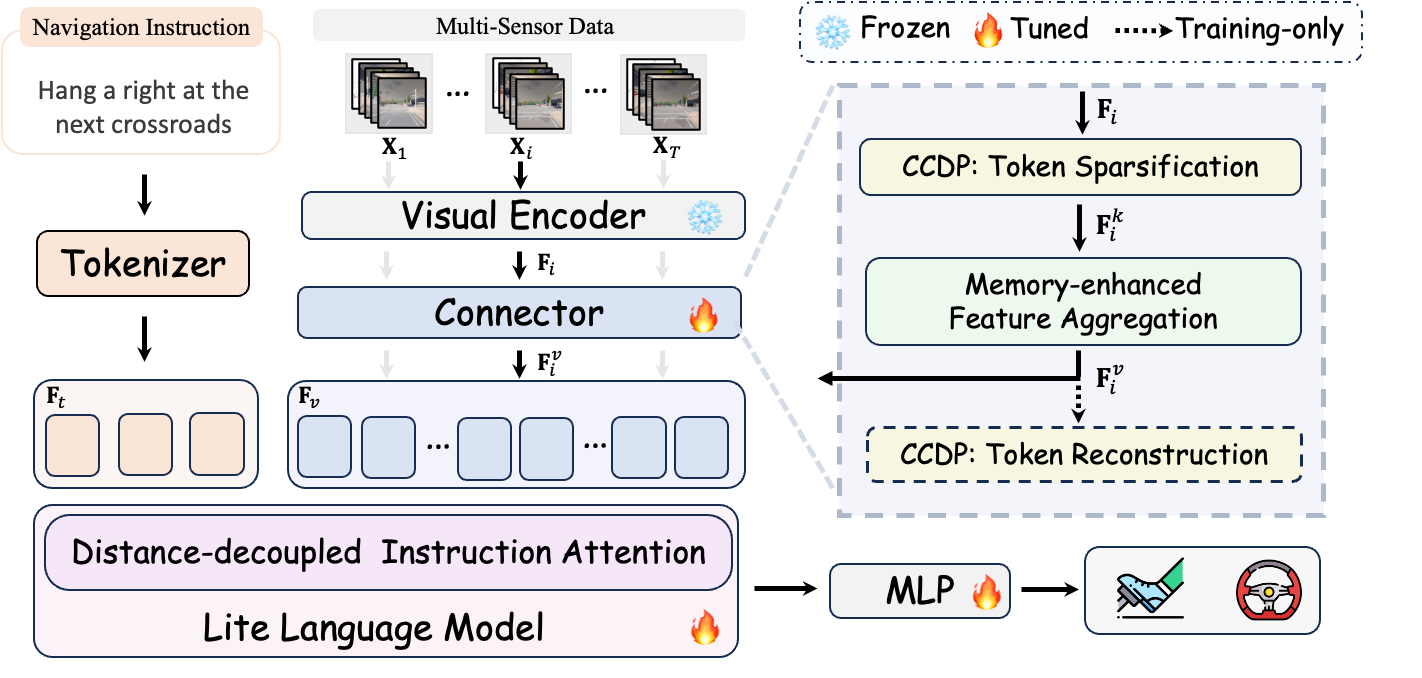}
\caption{An overview of our proposed VLDrive framework. Given a sequence of visual data, our connector transforms each frame's raw visual features $\mathbf{F}_i$ into sparse yet informative representations $\mathbf{F}_i^v$ through two key components: \textbf{CCDP: Token Sparsification} and \textbf{Memory-enhanced Feature Aggregation (MEFA)}. Subsequently, a lite language model augmented with \textbf{Distance-decoupled Instruction Attention (DDIA)} jointly processes tokenized navigation instructions $\mathbf{F}_t$ and temporal visual features $\mathbf{F}_v = \{\mathbf{F}_i^v \mid i = 1,\ldots,T\}$. The resulting hidden representations are fed into an MLP for trajectory prediction, followed by PID controllers that translate these predictions into concrete driving actions. Additionally, we incorporate \textbf{CCDP: Token Reconstruction} as a \textit{training-only} auxiliary task to further strengthen visual information integrity of $\mathbf{F}_i^v$ via explicit token reconstruction.}
\label{fig:overview}
\end{figure*}

\subsection{Token Reduction}
Transformer has achieved substantial performance improvements across a range of tasks, yet it faces criticism for its quadratic computational costs relative to the number of input tokens. To tackle this issue, numerous studies focus on reducing token counts~\cite{rao2021dynamicvit,kong2022spvit,xu2022evo,wei2023joint}. For instance, DynamicViT~\cite{rao2021dynamicvit} adaptively prunes redundant tokens at different Transformer layers to achieve token sparsification. SPVit~\cite{kong2022spvit} introduces an attention-based multi-head token selector with a soft pruning strategy. Unlike direct pruning methods, this approach consolidates selected redundant tokens into a single package token, enhancing efficiency while preserving essential information.
In the LLM era, token reduction is also a crucial strategy to reduce the heavy computation costs~\cite{li2023blip,zhu2023minigpt,li2023llama,shang2024llava}. Most existing works utilize Q-former~\cite{li2023blip} to aggregate visual features into a fixed number of tokens~\cite{li2023blip,zhu2023minigpt,zhang2023internlm}. LLaMa-VID~\cite{li2023llama} leverages average pooling followed by a learnable projector to compress the video frame representation. LLaVA-PruMerge~\cite{shang2024llava} proposes an adaptive token selection via outlier detection, paired with a cluster-based merging strategy to supplement the information of pruned tokens to kept ones. Distinct from prior approaches, we present a cycle-consistent dynamic pruning strategy that integrates token reconstruction with token sparsification at the training phase. This design not only enables more representative token selection but also ensures that the retained tokens aggregate fine-grained global visual information.

%% file: sec/3_method.tex
\section{Method}
\subsection{Overview}
This task aims to predict driving actions by leveraging sequences of multi-sensor data and corresponding human navigation instructions. As shown in Fig.~\ref{fig:overview}, our method follows the common framework of a multi-modal large language model (MLLM) and comprises three main components: 

\noindent \textbf{Visual Encoder}: Given a sequence of visual data $\{\mathbf{X}_i, i=1,2,...,T\}$, where each frame $\mathbf{X}_i$ includes multi-view camera images and corresponding LiDAR data. A visual encoder processes these multi-modal inputs to produce a unified feature representation $\mathbf{F}_i \in \mathbb{R}^{N \times C}$ for each frame, with $N$ representing the total number of tokens. 

\noindent\textbf{Connector}: As a bridge to align visual features with the language space, the connector plays a remarkable role in MLLM. To endow our model with adaptive visual perception and long-context modeling capability, as illustrated in Fig.~\ref{fig:ccdp}, we tailor cycle-consistent dynamic visual pruning (CCDP), consisting of token sparsification and reconstruction, and memory-enhanced feature aggregation (MEFA) for the connector, significantly enhancing the visual representation. We denote the improved features of each frame by the connector as $\mathbf{F}_i^v \in \mathbb{R}^{N_v \times C_t}$, where $C_t$ is the language model’s hidden dimension. 

\noindent\textbf{Lite Language Model}: A lightweight language model, enhanced by our proposed distance-decoupled instruction attention (DDIA), is utilized to process the combination of visual tokens of all frames, $\mathbf{F}_v \in \mathbb{R}^{N_v T \times C_t}$ and corresponding text embedding $\mathbf{F}_t \in \mathbb{R}^{N_t \times C_t}$ tokenized from navigation instructions. A following MLP leverages features from the language model and predicts the future trajectory, which is subsequently converted by PID controllers to produce the lateral steering action and the longitudinal acceleration action. We detail each component in the following subsections.     

\subsection{CCDP: Token Sparsification}
\label{sec:dp}
\begin{figure*}[t]
\centering
\includegraphics[width=0.85\textwidth, trim=20 0 0 0]{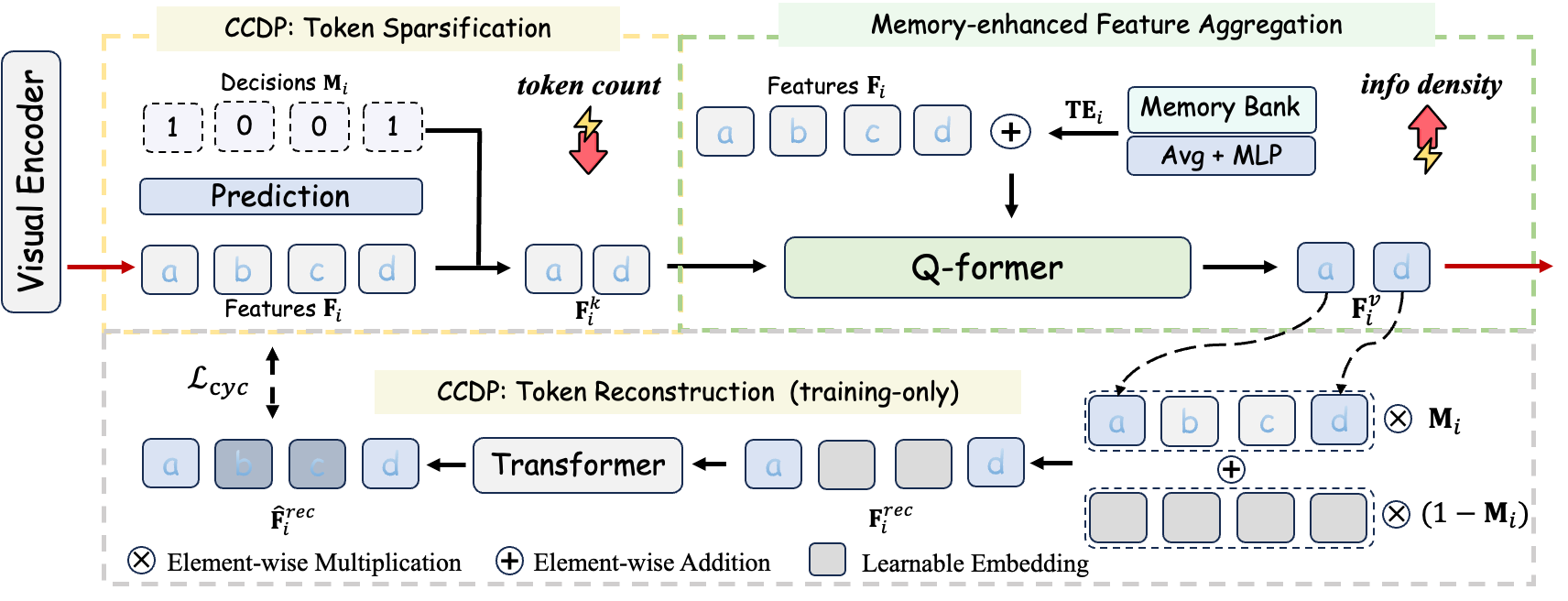}
\caption{A detailed illustration of our proposed connector. \textbf{CCDP: Token Sparsification} and \textbf{Memory-enhanced Feature Aggregation} are proposed to reduce token count while enhancing information density. \textbf{CCDP: Token Reconstruction} serves as a training-only task that further ensures the information integrity of the retained tokens. Red arrows (\textcolor{red}{$\rightarrow$}) indicate the input and output paths of our connector.}
\label{fig:ccdp}
\end{figure*}

Given the $i$-th frame's multi-modal features $\mathbf F_i$, as in DynamicViT~\cite{rao2021dynamicvit}, we predict the probabilities for pruning and retaining each token. Specifically, the probabilities are calculated based on the combined global and local features $[ \mathbf G_i;  \mathbf L_i]$, which are obtained by \text{MLP($\cdot)$} operations:
\begin{equation}
    \mathbf L_i = \text{MLP}( \mathbf{F}_i) \in\mathbb{R}^{N \times \frac{C}{2}}
\end{equation}
\begin{equation}
     \mathbf G_i = \text{Avg}( \mathbf L_i) \in\mathbb{R}^{1 \times \frac{C}{2}}
\end{equation}
where \text{Avg($\cdot$)} is average operation. Another MLP is utilized to predict the probability $\mathbf{S}_i$, and [$\cdot$;$\cdot$] denotes concatenation with broadcast mechanism:
\begin{equation}
     \mathbf{S}_i = \text{Softmax}(\text{MLP}([ \mathbf{G}_i; \mathbf{L}_i])) \in\mathbb{R}^{N \times 2}
\end{equation}
In our implementation, we leverage Gumbel-Softmax to obtain the binary mask:
\begin{equation}
   \mathbf{M}_i = \text{Gumbel-Softmax}(\mathbf{S}_i)_{*,1} \in \{0,1\}^N 
\end{equation}
where $\mathbf{M}_i$ denotes the token retention masks. As shown in Fig.~\ref{fig:ccdp}, the kept tokens $\mathbf{F}_i^k$ selected based on $\mathbf{M}_i$ are delivered to the \textbf{Memory-enhanced Feature Aggregation} module, which is detailed in the next subsection. 

\subsection{Memory-enhanced Feature Aggregation}
\label{sec:memory}
Temporal reasoning is integral to our everyday driving scenarios. Human drivers can deduce the likely future direction or intentions of an approaching vehicle by analyzing its historical trajectory, and subsequently tailor their driving actions. To attain autonomous driving that matches the safety and reliability of human driving, we explicitly inject the temporal context into the current visual tokens. Specifically, for the current time step $i$, we introduce a memory bank storing the adjacent $Z$ frames from the current one, denoted as $\mathbf {B}_i = [\mathbf F_{i-Z},\mathbf F_{i-Z+1},..., \mathbf F_{i-1}]$, $\mathbf{B}_i \in\mathbb{R}^{Z \times N \times C}$. Temporal encoding $\mathbf{TE}_i$ is derived by an MLP utilizing the historical average data and current frame information:
\begin{equation}
   \mathbf B_i^{avg} = \text{Avg}( \mathbf B_i) \in\mathbb{R}^{N \times C}
\end{equation}
\begin{equation}
   \mathbf{TE}_i = \text{MLP}([\mathbf F_i; \mathbf B_i^{avg}])
\end{equation}

Afterwards, a query transformer (Q-former)~\cite{li2023blip} is introduced to aggregate the features of current frames enhanced by temporary encoding:
\begin{equation}
   \mathbf{F}_i^v = \text{Q-Former}(\mathbf F_i^k, \mathbf F_i+\mathbf {TE}_i)
\end{equation}

In contrast to the original Q-former, we implement two key modifications: 1) Instead of using learnable queries, we utilize the retained visual tokens $\mathbf F_i^k$ as queries to probe and aggregate features. These retained tokens serve as significant anchors within the current frame, allowing for more effective feature aggregation. 2) We introduce a temporary embedding $\mathbf{TE}_i$ to the original visual features $\mathbf{F}_i$, which enhances the model's ability to focus on temporary-salient moving objects.

Finally, the generated $\mathbf F_i^v$ serves as part of the aggregated visual tokens across all frames, $\mathbf{F}_v$, which is then fed into the subsequent language model for trajectory prediction based on the corresponding navigation instructions. Additionally, we incorporate \textbf{CCDP: Token Reconstruction} as a \textit{training-only} auxiliary task to further strengthen visual information integrity via explicit token reconstruction.

\subsection{CCDP: Token Reconstruction}
Through memory-enhanced feature aggregation, the generated $\mathbf {F}_{i}^v $ ideally encapsulates all visual features of the current frame. To further enhance the connector's ability to select critical visual tokens and aggregate complete information, we expect to reconstruct features of those pruned tokens based on $\mathbf {F}_{i}^v $. Firstly, we add the enhanced tokens $\mathbf {F}_{i}^v$ into the original features $\mathbf{F}_{i}$ according to their original token positions, forming a new feature vector denoted as $\langle \text{MLP}(\mathbf{F}_{i}^v), \mathbf{F}_{i}\rangle$, where an MLP function is used to align the dimensions of $\mathbf{F}_{i}^{v}$ and $\mathbf{F}_{i}$. Secondly, drawing inspiration from the Masked Autoencoder~\cite{he2022masked}, a learnable embedding $\mathbf e \in \mathbb{R}^{N \times C}$ is introduced as a foundation for reconstructing the pruned tokens. Finally, we construct the input feature vector to be reconstructed through the following formula:
\begin{equation}
    \mathbf{F}^{rec}_{i} = 
        \langle \text{MLP} (\mathbf {F}_{i}^v), \mathbf {F}_{i}\rangle \cdot \mathbf{M}_{i} + 
        \mathbf e \cdot (1-\mathbf{M}_{i})
\end{equation}
In this way, we incorporate the decision $\mathbf{M}_{i}$ into the computational graph of feature reconstruction, enabling the reconstruction loss to influence pruning decisions and compelling the model to retain the most critical tokens. Afterwards, a 4-layer Transformer block takes as input $\mathbf{F}_i^{rec}$ and predicts the reconstructed results $ \mathbf{\hat F}_i^{rec}$. A cycle consistency constraint (Eq.~\ref{eq:loss_cyc}) is imposed to ensure that the retained tokens $\mathbf{F}_{i}^{v}$ preserve the integral information of the current video frame.

\subsection{Distance-decoupled Instruction Attention}
\label{sec:att}
\begin{figure}[t]
\centering
\includegraphics[width=0.56\linewidth, trim=0 0 0 0]{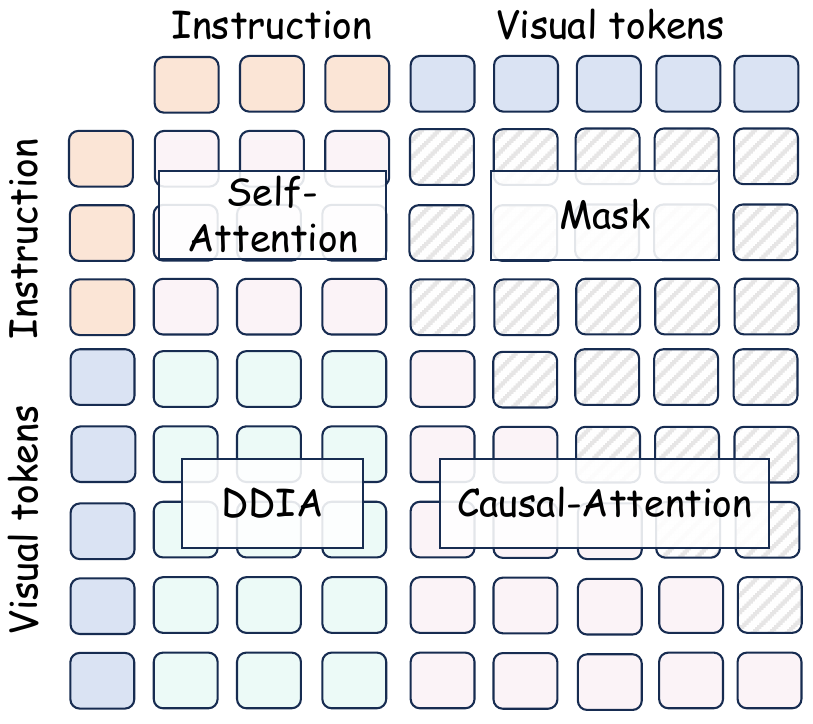}
\caption{A detailed illustration of our proposed Distance-decoupled Instruction Attention (DDIA).}
\label{fig:ddia}
\end{figure}

We introduce a \textbf{lightweight language model} for efficient autonomous driving, reducing both the number of parameters and computational resource requirements.
Furthermore, distance-decoupled instruction attention (DDIA) is tailored for our lite language model, significantly alleviating the attention dilution for navigation instructions in the long-context driving scenarios. Specifically, as shown in Fig.~\ref{fig:ddia}, our proposed DDIA diverges from the traditional vanilla attention in two key aspects:

\noindent 1): Causal-attention $\boldsymbol{\rightarrow}$ Self-attention: Rather than adhering to the original causal setting, our self-attention \textbf{within instructional segments} allows each text token to access ample contextual information, thereby producing a more comprehensive representation of text features.

\noindent 2): Distance-dependent $\boldsymbol{\rightarrow}$ Distance-decoupled:
In contemporary research on LLM, position encoding (e.g. RoPE~\cite{su2024roformer}) is recognized as a critical ingredient to explicitly inject positional information into input tokens and enhance the model's contextual modeling capabilities.
However, position encodings may introduce challenges, particularly in long-range driving scenarios. As historical visual data accumulates, the distance between current visual tokens and instructional tokens increases. The feature disparity introduced by position encoding may negatively impact the attention sensitivity of visual tokens to instruction ones, resulting in trajectory predictions that deviate from the specified instructions. To address this problem, inspired by \cite{ma2023vista}, we maintain the distance-dependent attention within the unimodal text or visual tokens, but make the cross-modal attention between instructions and visual tokens unaffected by the position encoding. 
Specifically, with $\mathcal{V}$ and $\mathcal{I}$ denoting the sets of visual and instruction tokens, taking the $j$-th position as an example, our DDIA is defined as:
\begin{equation}
\scriptsize
\begin{split}
 \text{DDIA}(\mathbf{Q}, \mathbf{K}, \mathbf{V})_{j} = \frac{\sum\limits_{k_i \in \mathcal{I}}sim(R({\mathbf{q}}_{j}), R({\mathbf{k}}_{i}))\mathbf{v}_{i} }{\sum\limits_{k_i \in \mathcal{I}}sim(R({\mathbf{q}}_{j}), R({\mathbf{k}}_{i}))}, \text{if~} \mathbf{q}_{j} \in \mathcal{I}, \\
 = \frac{\sum\limits_{k_i \in \mathcal{I}}sim({\mathbf{q}}_{j}, {\mathbf{k}}_{i})\mathbf{v}_{i} + \sum\limits_{k_{i} \in \{\mathcal{V} \mid < j\}}sim(R(\mathbf{q}_{j}), R(\mathbf{k}_{i}))\mathbf{v}_{i}}{\sum\limits_{k_i \in \mathcal{I}}sim({\mathbf{q}}_{j}, {\mathbf{k}}_{i}) + \sum\limits_{k_{i} \in \{\mathcal{V} \mid < j\}}sim(R(\mathbf{q}_{j}), R(\mathbf{k}_{i}))}, \text{if~} \mathbf{q}_{j} \in \mathcal{V} 
\end{split}
\label{eq:ddia-t}
\end{equation}
where $R(\cdot)$ denotes RoPE position embedding, $sim(\mathbf{q}_{j}, \mathbf{k}_{i}) = exp(\mathbf{q}_{j}^{T}\mathbf{k}_{i}/\sqrt{d})$ and $\{\mathcal{V} \mid < j\}$ represents the causal subset within visual tokens.

\subsection{Objective Function}
\label{sec:loss}
The full training objective for $i$-th frame consists of two parts, formulated as follows:
\begin{equation}
\mathcal{L}= {\mathcal{L}_{way}} + {\mathcal{L}_{cyc}}, 
\label{eq:loss}
\end{equation}
here, $\mathcal{L}_{way}$ supervises trajectory prediction by minimizing the $L_1$ norm between predicted and ground-truth paths~\cite{shao2023lmdrive}, while ${\mathcal{L}_{cyc}}$ facilitates our model to select the most discriminative tokens and aggregate more representative visual features, which is composed of two adversarial losses: 
\begin{equation}
\mathcal{L}_{cyc}= \lambda_1{\mathcal{L}_{prun}} + \lambda_2{\mathcal{L}_{rec}} 
\label{eq:loss_cyc}
\end{equation}
\begin{equation}
\mathcal{L}_{prun} = (R-\frac{1}{N}\sum_{j=1}^N \mathbf{M}_{ij})^2 
\label{eq:loss_prun}
\end{equation}
\begin{equation}
\mathcal{L}_{rec} = \sum_{j=1}^N(1-\mathbf{M}_{ij}) (\mathbf{F}_{ij} - \mathbf{\hat F}_{ij}^{rec})^2
\label{eq:loss_rec}
\end{equation}
${\mathcal{L}_{prun}}$ restricts the ratio of kept tokens to our pre-defined value $R$, and ${\mathcal{L}_{rec}}$ reconstructs pruned tokens from kept ones after memory-enhanced feature aggregation. $\lambda_1$ and $\lambda_2$ balance the weights of two loss items.
$ \mathbf{M}_{ij}$, $\mathbf{F}_{ij}$ and $\mathbf{\hat F}_{ij}^{rec}$ denote the $j$-th element of $\mathbf M_i$, $\mathbf {F}_{i}$ and $\mathbf{\hat F}_{i}^{rec}$, respectively.

%% file: sec/4_experiments.tex
\begin{table*}[t]
\centering
\caption{Performance comparison of our method with LMDrive incorporated various LLM backbones on the LangAuto benchmark. We present the metrics from 3 evaluation runs.}
\resizebox{0.75\textwidth}{!}{
\begin{tabular}{cccccc}
\toprule
\multirow{2}{*}{Method}  &   \multirow{2}{*}{LLM} &    \multirow{2}{*}{Params} & \multicolumn{3}{c}{LangAuto} \\ \cmidrule(r){4-6} 
                       & &  (Total)&  DS $\uparrow$  & RC $\uparrow$   &  IS $\uparrow$    \\ \cmidrule(r){1-3} \cmidrule(r){4-6} 
\multirow{9}{*}{LMDrive}  &  \multicolumn{5}{c}{\textit{\cellcolor[gray]{0.9}Large Language Model}}  \\
                  &  LLaMA~\cite{touvron2023llama1} & 7B     & 31.3$\pm$1.5   &  37.1$\pm$1.6  & 0.82$\pm$0.01   \\ 
                  &   LLaMA2~\cite{touvron2023llama2}  & 7B     & 32.8$\pm$2.1   &  40.1$\pm$2.2  & 0.81$\pm$0.02     \\ 
                  &  Vicuna~\cite{zheng2023judging}  & 7B    &  33.5$\pm$1.9  &  39.3$\pm$1.9  &  0.83$\pm$0.02    \\ 
                  &  Vicuna-v1.5~\cite{zheng2023judging}  & 7B    &  34.0$\pm$3.8  &  39.0$\pm$3.3  &  \textbf{{0.85}$\pm${0.06}}   \\ 
                  &   LLaVA-v1.5~\cite{liu2023improved} & 7B   &  {36.2}$\pm${2.3}  &  {46.5}$\pm${4.3}  &  0.81$\pm$0.03    \\ 
                  & \multicolumn{5}{c}{\textit{\cellcolor[gray]{0.9}Lightweight Language Model}} \\ 
                  &   Phi-2~\cite{javaheripi2023phi} & 3B   &  {22.3}$\pm${0.1}  &  {28.9}$\pm${1.0}  &  {0.82}$\pm${0.02}  \\ 
                  &   TinyLLaMA~\cite{zhang2024tinyllama} & 1.3B   &  {25.2}$\pm${2.3}  &  {38.6}$\pm${3.7}  &  {0.71}$\pm${0.02}   \\ 
                  \midrule
\multirow{2}{*}{Ours}  & LLaMA* &  1.3B & 41.7$\pm$1.8   &  52.6$\pm$2.5   & 0.81$\pm$0.02  \\
 & TinyLLaMA~\cite{zhang2024tinyllama} &  1.3B &  \textbf{43.8$\pm$2.4} & \textbf{54.5$\pm$3.0}   &0.84$\pm$0.02 \\
\bottomrule                    
\end{tabular}}
\label{table:compare_long}
\end{table*}

\begin{table*}[t]
\centering
\caption{Performance comparison of our method with LMDrive incorporated various LLM backbones on the LangAuto-Short and LangAuto-Tiny benchmarks. We present the metrics from 3 evaluation runs.}
\resizebox{\textwidth}{!}{
\begin{tabular}{ccccccccc}
\toprule
\multirow{2}{*}{Method}  & \multirow{2}{*}{LLM } &\multirow{2}{*}{Params} & \multicolumn{3}{c}{LangAuto-Short} & \multicolumn{3}{c}{LangAuto-Tiny} \\ 
\cmidrule(r){4-6} \cmidrule(r){7-9} 
                       & &(Total)&  DS $\uparrow$  & RC $\uparrow$   &  IS $\uparrow$      &  DS $\uparrow$  & RC $\uparrow$   &  IS $\uparrow$   \\ \cmidrule(r){1-3} \cmidrule(r){4-6} \cmidrule(r){7-9} 
\multirow{6}{*}{LMDrive} &     \multicolumn{8}{c}{\textit{\cellcolor[gray]{0.9}Large Language Model}}   \\
                  & LLaMA~\cite{touvron2023llama1} & 7B     & 42.8$\pm$7.2     &  49.1$\pm$8.5    &  0.87$\pm$0.03   &  52.2$\pm$5.3    &  57.8$\pm$8.0    &  0.91$\pm$0.05   \\ 
                  &  LLaMA2~\cite{touvron2023llama2}    & 7B   &  44.8$\pm$6.2    &  53.5$\pm$5.5    &  0.84$\pm$0.02   &  56.1$\pm$4.1    &  64.2$\pm$4.7    &  0.87$\pm$0.04    \\ 
                  &  Vicuna~\cite{zheng2023judging}   & 7B &    45.3$\pm$4.9  &  54.3$\pm$3.9    &  0.83$\pm$0.03 &  55.5$\pm$3.9    &  63.1$\pm$4.2    &  0.88$\pm$0.04    \\ 
                  &  Vicuna-v1.5~\cite{zheng2023judging}  & 7B   &   47.0$\pm$4.3   &  56.5$\pm$2.4    &  0.83$\pm$0.04    &  59.0$\pm$2.6    &  69.9$\pm$2.3    &  0.84$\pm$0.02    \\ 
                  &  LLaVA-v1.5~\cite{liu2023improved}   & 7B &   50.6$\pm$1.7   &  60.0$\pm$3.4    &  0.84$\pm$0.04   &  66.5$\pm$3.6    &  77.9$\pm$2.3    &  0.85$\pm$0.02    \\ 
                  & \multicolumn{8}{c}{\textit{\cellcolor[gray]{0.9}Lightweight Language Model}} \\ 
                   &   Phi-2~\cite{javaheripi2023phi} & 3B   &  {48.0}$\pm${3.6}  &  {55.2}$\pm${4.1}  &  \textbf{{0.87}$\pm${0.01}} &  {56.3}$\pm${4.9}  &  {68.2}$\pm${3.9}  &  {0.82}$\pm${0.04}  \\ 
                  &   TinyLLaMA~\cite{zhang2024tinyllama} & 1.3B   &  {46.2}$\pm${3.4}  &  {59.7}$\pm${3.4}  &  {0.79}$\pm${0.02}  &   {64.1}$\pm${1.2}  &  {75.0}$\pm${1.4}  &  {0.86 }$\pm${0.01}    \\ 
                  \midrule
\multirow{2}{*}{Ours} & LLaMA* &   1.3B &  63.6$\pm$4.1  &  77.6$\pm$4.1 & 0.81$\pm$0.06 & 78.4$\pm$4.6 &\textbf{85.6$\pm$2.4} & 0.91$\pm$0.04 \\
& TinyLLaMA~\cite{zhang2024tinyllama} &   1.3B & \textbf{67.4$\pm$2.0}  & \textbf{78.1$\pm$2.3}  &0.85$\pm$0.02  &  \textbf{81.9$\pm$0.9} & 85.5$\pm$0.6& \textbf{0.94$\pm$0.02} \\
                  \bottomrule
\end{tabular}}

\label{table:compare_tiny_short}
\end{table*}

\section{Experiments}
\subsection{Experiment Settings}
\paragraph{Datasets.} We train our model on the official language-driven autonomous driving dataset~\cite{shao2023lmdrive}, which includes 64K instruction-following data clips collected across 8 towns on CARLA~\cite{dosovitskiy2017carla} simulation environment. Each clip not only equips multi-sensor input data (multi-view camera images and LiDAR data), but also introduces aligned navigation instructions in the natural language format. 

\paragraph{Implementation Details.} We adopt the visual encoder in LMDrive~\cite{shao2023lmdrive}, producing 106 visual tokens ($N$=106) and kept it fixed during training. Two configurations of lightweight language models are integrated into our framework: LLaMA* (a 4-layer lite version of LLaMA~\cite{touvron2023llama1}) and TinyLLaMA~\cite{zhang2024tinyllama}.
The hype parameters $R$ determining the ratio of kept tokens and $Z$ denoting the number of stored adjacent frames in the memory bank are set to $0.3$ and $10$, respectively. $\lambda_1$ and $\lambda_2$ in the loss function (Eq.~\ref{eq:loss_cyc}) are configured to 10 and 1. More details are provided in the \textit{Supplementary Material}.

\paragraph{Evaluation Metrics.}
As in LMDrive~\cite{shao2023lmdrive}, we evaluate our model on three benchmarks, categorized by route length: LangAuto ($>$500m), LangAuto-Short (150m$–$500m) and LangAuto-Tiny ($<$150m) and employ route completion (RC), infraction score (IS), and driving score (DS) to assess close-looped driving performance. Specifically, route completion measures the extent to which a specified route is completed. Infraction score quantifies driving safety, starting from an ideal 1.0 base score and decrementing proportionally in the event of collisions or traffic rule violations. Driving Score is calculated as the product of RC and IS, offering a comprehensive evaluation of overall performance.

\subsection{Comparisons with LLM-based Agents}
We compare our VLDrive with LMDrive, a pioneering LLM-driven, closed-loop autonomous driving method, across several distance benchmarks. Table~\ref{table:compare_long} presents the comparison results on the LangAuto benchmark. VLDrive has approximately 81\% fewer parameters yet demonstrates superior driving performance compared to LMDrive, which is equipped with various LLM configurations. Specifically, VLDrive-LLaMA* achieves driving scores of 41.7\% and route completion rates of 52.6\%, outperforming LMDrive with LLaVA-v1.5 by 5.5\% and 6.1\%. This trend continues on the LangAuto-Short and LangAuto-Tiny benchmarks, as shown in Table~\ref{table:compare_tiny_short}, where VLDrive-LLaMA* attains driving scores of 63.6\% and 78.4\%, surpassing LMDrive by 13\% and 11.9\%. VLDrive-TinyLLaMA achieves new state-of-the-art performance, boosting the DS score to 43.8\%, 67.4\%, and 81.9\% on the LangAuto, LangAuto-Short, and LangAuto-Tiny benchmarks, respectively.

\subsection{Ablation Study}
To investigate the impact of various components of our method, we conduct comprehensive ablation studies on the standard LangAuto benchmark. To reduce time costs and enhance training efficiency, we randomly extract approximately 25\% of the training data from each town, forming a mini-training set for all ablation experiments. We present the metrics from 3 evaluation runs.

\begin{figure}[t]
\centering
\includegraphics[width=0.88\linewidth, trim=0 0 0 0]{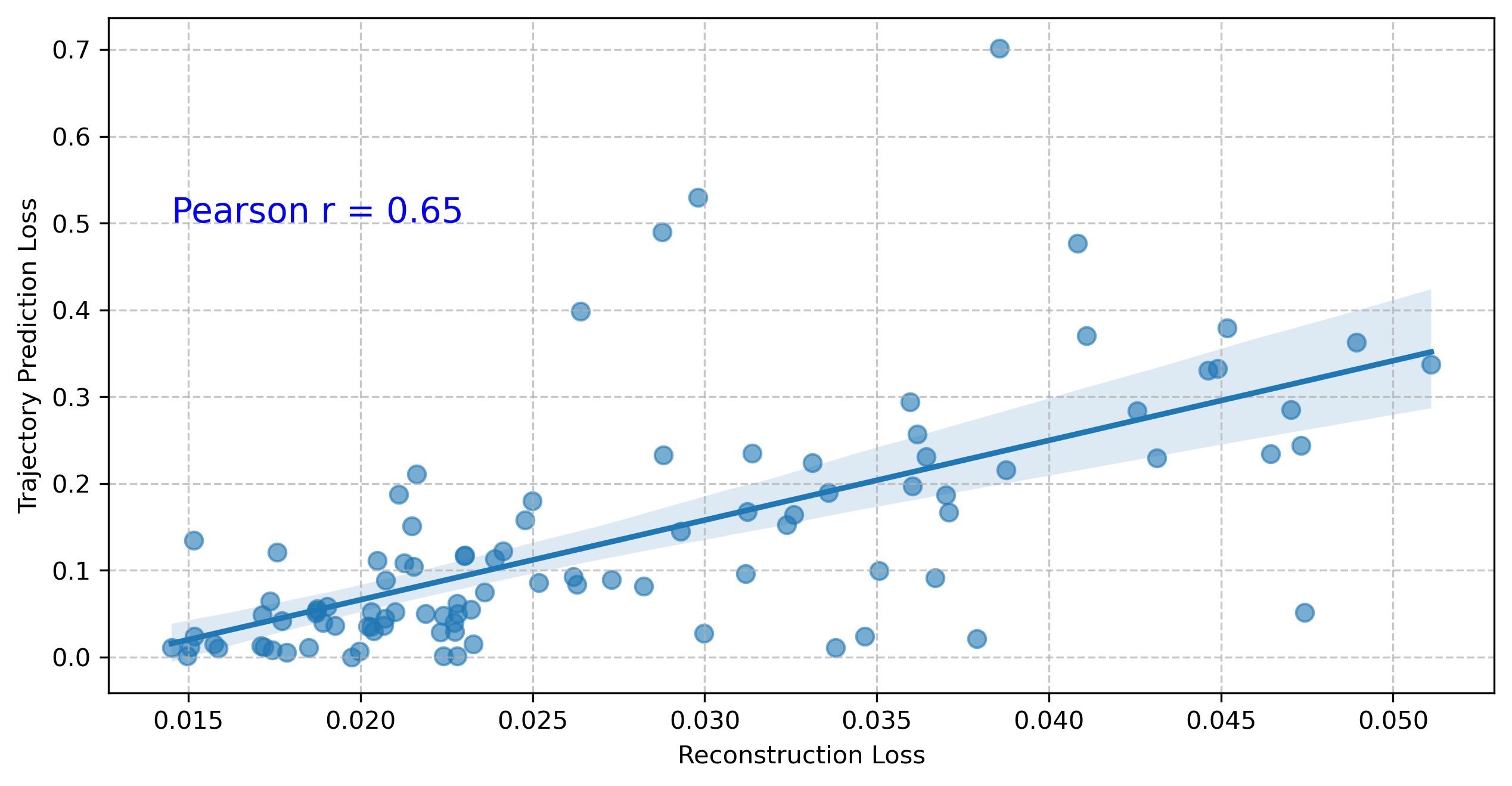}
\caption{Correlation analysis between reconstruction and trajectory prediction losses, revealing a significant positive relationship (Pearson's r = 0.65).}
\label{fig:loss}
\end{figure}

\paragraph{Effectiveness of Each Individual Component.} As shown in Table~\ref{tab:each_component}, we begin our ablation study from the baseline, which utilizes LLaMA* as the trainable language model and Q-Former~\cite{li2023blip} with learnable queries as the connector. Our method improves the baseline by: (1) introducing CCDP to extract critical sparse tokens, (2) replacing the vanilla Q-Former with our proposed MEFA module, and (3) substituting the causal attention mechanism with our DDIA module. The results in Table~\ref{tab:each_component} validate the contribution of each component, with their synergistic integration leading to substantial improvements in driving performance.
\paragraph{Effectiveness of CCDP.}
We compare our CCDP approach with other token reduction strategies, including: 1) Structural pooling, a token reduction technique based on average pooling operations. 2) Dynamic pruning without token reconstruction. We maintain a consistent token retention ratio across all methods. 
As shown in Table~\ref{tab:token_reduction}, benefiting from the cycle-consistency constraints, CCDP is able to extract the most critical visual information and consequently achieves the highest driving scores. Fig.~\ref{fig:loss} further illustrates the positive correlation between reconstruction loss and trajectory prediction loss on the open-looped validation set, with a pearson correlation coefficient of 0.65.
\paragraph{Analysis of Token Sparsification.} Table~\ref{tab:token_ratio} demonstrates that, retaining 30\% of the visual tokens captures critical clues for autonomous driving. Further increment in the number of tokens produces no significant performance gains in driving scores.

\begin{table}[t]
    \centering                
    \caption{Ablation studies of each individual component.}
    \resizebox{0.9\linewidth}{!}{
      \begin{tabular}[t]{@{}l|ccc@{}}
        \toprule
        Method  &  DS $\uparrow$  & RC $\uparrow$   &  IS $\uparrow$ \\
        \midrule
        Baseline & 30.0$\pm$0.8 &46.2$\pm$3.5 & 0.69$\pm$0.03  \\
        \hline
        \rowcolor[gray]{.9} Ours  &39.8$\pm$1.7 &49.1$\pm$1.8& 0.83$\pm$0.02 \\
        w/o CCDP  & 35.5$\pm$3.8 &45.3$\pm$3.5&0.78$\pm$0.02 \\
        w/o MEFA &35.9$\pm$0.8&46.9$\pm$1.3&0.81$\pm$0.02 \\
        w/o DDIA & 36.3$\pm$1.3&47.0$\pm$1.0&0.81$\pm$0.02 \\
        \bottomrule
      \end{tabular}}
      \label{tab:each_component} 
\end{table}

\begin{table}[t]
    \centering               
    \caption{Ablation studies of various token reduction strategies.}
    \resizebox{0.92\linewidth}{!}{
      \begin{tabular}[t]{@{}l|ccc@{}}
        \toprule
        Method  &  DS $\uparrow$  & RC $\uparrow$   &  IS $\uparrow$ \\
        \midrule
        Structural Pooling &33.6$\pm$2.4 & 45.8$\pm$4.3&0.78$\pm$0.06\\
        Dynamic Pruning  &36.3$\pm$2.6&48.0$\pm$2.3 & 0.76$\pm$0.03\\
        \rowcolor[gray]{.9} CCDP~(Ours)    &39.8$\pm$1.7 &49.1$\pm$1.8& 0.83$\pm$0.02\\
        \bottomrule
      \end{tabular}}
      \label{tab:token_reduction} 
    \end{table}  

    \begin{table}[t]
    \centering                
    \caption{Ablation studies of different token retention ratios.}
    \resizebox{0.92\linewidth}{!}{
      \begin{tabular}[t]{@{}c|ccc@{}}
        \toprule
         Retention Ratio ($R$)  &  DS $\uparrow$  & RC $\uparrow$   &  IS $\uparrow$ \\
        \midrule
        5\% & 33.4$\pm$0.4&45.1$\pm$3.2& 0.77$\pm$0.03\\
        10\% &35.6$\pm$1.7&47.3$\pm$1.0& 0.79$\pm$0.02\\
        \rowcolor[gray]{.9} 30\%  &39.8$\pm$1.7 &49.1$\pm$1.8& 0.83$\pm$0.02\\
        50\%    & 39.8$\pm$1.2&51.4$\pm$3.7&0.79$\pm$0.04 \\
        \bottomrule
      \end{tabular}}
      \label{tab:token_ratio} 
    \end{table}

    \begin{table}[t]
    \centering
     \caption{Ablation of memory bank with different capacity.}
    \resizebox{0.92\linewidth}{!}{
    \begin{tabular}[t]{@{}c|ccc@{}}
                \toprule
                 Capacity ($Z$) &  DS $\uparrow$  & RC $\uparrow$   &  IS $\uparrow$ \\
                \midrule
                 5 &37.1$\pm$1.9&49.1$\pm$0.5& 0.80$\pm$0.01\\
                 \rowcolor[gray]{.9}  10 &39.8$\pm$1.7 &49.1$\pm$1.8& 0.83$\pm$0.02 \\
                 20 &38.3$\pm$3.0 &49.9$\pm$2.6& 0.81$\pm$0.02\\
                \bottomrule
          \end{tabular}}
          \label{tab:memory_bank}
        \end{table}

\paragraph{Effectiveness of MEFA.} As illustrated in Table~\ref{tab:memory_bank}, incorporating a memory bank provides essential temporal information that aids in predicting the future trajectories and behaviors of the ego-car and other objects, thereby enhancing the reliability of autonomous driving. Furthermore, our experiments indicate that a capacity of 10 adjacent frames yields the most significant performance improvements.

\begin{figure}[t]
\centering
\includegraphics[width=0.85\linewidth, trim=0 0 0 0]{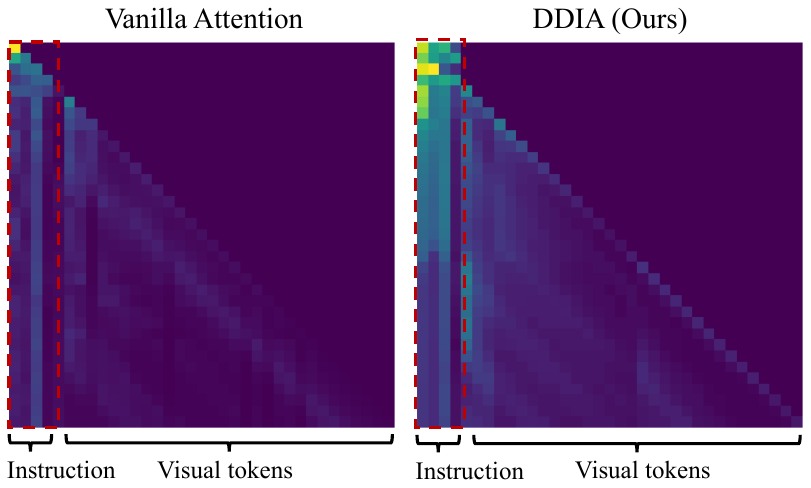}
\caption{Visual comparison of attention maps generated by the language model equipped with vanilla causal attention and DDIA. We aggregate the attention weights within the instruction segment into the first four tokens for clearer visualization.}
\label{fig:atts}
\vspace{-1em}
\end{figure}

\paragraph{Effectiveness of DDIA.} Fig.~\ref{fig:atts} shows the visual comparison of attention maps generated by vanilla causal attention and DDIA, where brighter colors indicate higher values. As highlighted in the red dashed box, DDIA enhances attention to instruction tokens, enabling visual tokens to better integrate instructional information and thus make more accurate driving decisions.

%% file: sec/5_conclusion.tex
\section{Conclusion}
In this work, we introduce VLDrive, a vision-enhanced, lightweight model for language-grounded autonomous driving. To facilitate the practical deployment of autonomous vehicles, we streamline the heavy LLM into a more compact form while augmenting its visual modeling and vision-language alignment. Specifically, we employ cycle-consistent visual dynamic pruning to efficiently capture the most salient visual information for each frame, and incorporate a memory-enhanced feature aggregation that enriches critical temporal information to help the model comprehend historical trajectories and infer future movements. Additionally, a distance-decoupled instruction attention strategy is tailored for our lightweight language model, alleviating the attention dilution to navigation instructions and boosting the model's instruction-following capability. Extensive experiments demonstrate that our model, despite having significantly fewer parameters, achieves superior and robust driving performance across various benchmarks. 